%% file: ijcai_2019_main.tex
\documentclass[english]{article}
\usepackage[T1]{fontenc}
\usepackage[latin9]{inputenc}
\usepackage{color}
\usepackage{babel}
\usepackage{float}
\usepackage{amsmath}
\usepackage{amssymb}
\usepackage{graphicx}
\usepackage{xargs}[2008/03/08]
\usepackage[unicode=true,pdfusetitle,
 bookmarks=true,bookmarksnumbered=false,bookmarksopen=false,
 breaklinks=false,pdfborder={0 0 1},backref=false,colorlinks=false]
 {hyperref}

\makeatletter

\floatstyle{ruled}
\newfloat{algorithm}{tbp}{loa}
\providecommand{\algorithmname}{Algorithm}
\floatname{algorithm}{\protect\algorithmname}

\usepackage{algolyx}
\usepackage{algolyx}

\usepackage{ijcai18}

\usepackage{times}

\usepackage{mathptmx}

\usepackage[small]{caption}
\usepackage{graphicx}

\usepackage{colortbl} 
\definecolor{header_color}{rgb}{0.74,0.88,0.91}
\definecolor{even_color}{rgb}{0.9,0.9,0.9}
\definecolor{subheader_color}{rgb}{0.85,0.93,0.95}
\definecolor{childheader_color}{rgb}{1.0,0.93,0.87}

\makeatother

\begin{document}

\title{Practical Batch Bayesian Optimization for Less Expensive Functions}

\author{Vu Nguyen, Sunil Gupta, Santu Rana, Cheng Li, Svetha Venkatesh \\
Deakin University, Australia\\
\{v.nguyen,sunil.gupta,santu.rana,cheng.l,svetha.venkatesh\}@deakin.edu.au}

\maketitle
\input{macros.tex}

\begin{abstract}
Bayesian optimization (BO) and its batch extensions are successful
for optimizing expensive black-box functions. However, these traditional
BO approaches are not yet ideal for optimizing less expensive functions
when the computational cost of BO can dominate the cost of evaluating
the black-box function. Examples of these less expensive functions
are cheap machine learning models, inexpensive physical experiment
through simulators, and acquisition function optimization in Bayesian
optimization. In this paper, we consider a batch BO setting for situations
where function evaluations are less expensive. Our model is based
on a new exploration strategy using geometric distance that provides
an alternative way for exploration, selecting a point far from the
observed locations. Using that intuition, we propose to use Sobol
sequence to guide exploration that will get rid of running multiple
global optimization steps as used in previous works.  Based on
the proposed distance exploration, we present an efficient batch BO
approach. We demonstrate that our approach outperforms other baselines
and global optimization methods when the function evaluations are
less expensive.
\end{abstract}

\section{Introduction}

\input{Introduction.tex}

\section{Batch BO for Less Expensive Functions}

\input{framework_v2.tex}

\section{Experiments}

\input{Experiments_v1.tex}

\section{Conclusion }

\input{Conclusion.tex}

\pagebreak{}

\bibliographystyle{named}
\bibliography{vunguyen}

\end{document}

%% file: macros.tex
\newcommand{\sidenote}[1]{\marginpar{\small \emph{\color{Medium}#1}}}

\global\long\def\se{\hat{\text{se}}}

\global\long\def\interior{\text{int}}

\global\long\def\boundary{\text{bd}}

\global\long\def\ML{\textsf{ML}}

\global\long\def\GML{\mathsf{GML}}

\global\long\def\HMM{\mathsf{HMM}}

\global\long\def\support{\text{supp}}

\global\long\def\new{\text{*}}

\global\long\def\stir{\text{Stirl}}

\global\long\def\mA{\mathcal{A}}

\global\long\def\mB{\mathcal{B}}

\global\long\def\mF{\mathcal{F}}

\global\long\def\mK{\mathcal{K}}

\global\long\def\mH{\mathcal{H}}

\global\long\def\mX{\mathcal{X}}

\global\long\def\mZ{\mathcal{Z}}

\global\long\def\mS{\mathcal{S}}

\global\long\def\Ical{\mathcal{I}}

\global\long\def\mT{\mathcal{T}}

\global\long\def\Pcal{\mathcal{P}}

\global\long\def\dist{d}

\global\long\def\HX{\entro\left(X\right)}
 \global\long\def\entropyX{\HX}

\global\long\def\HY{\entro\left(Y\right)}
 \global\long\def\entropyY{\HY}

\global\long\def\HXY{\entro\left(X,Y\right)}
 \global\long\def\entropyXY{\HXY}

\global\long\def\mutualXY{\mutual\left(X;Y\right)}
 \global\long\def\mutinfoXY{\mutualXY}

\global\long\def\given{\mid}

\global\long\def\gv{\given}

\global\long\def\goto{\rightarrow}

\global\long\def\asgoto{\stackrel{a.s.}{\longrightarrow}}

\global\long\def\pgoto{\stackrel{p}{\longrightarrow}}

\global\long\def\dgoto{\stackrel{d}{\longrightarrow}}

\global\long\def\lik{\mathcal{L}}

\global\long\def\logll{\mathit{l}}

\global\long\def\vectorize#1{\mathbf{#1}}

\global\long\def\vt#1{\mathbf{#1}}

\global\long\def\gvt#1{\boldsymbol{#1}}

\global\long\def\idp{\ \bot\negthickspace\negthickspace\bot\ }
 \global\long\def\cdp{\idp}

\global\long\def\das{}

\global\long\def\id{\mathbb{I}}

\global\long\def\idarg#1#2{\id\left\{  #1,#2\right\}  }

\global\long\def\iid{\stackrel{\text{iid}}{\sim}}

\global\long\def\bzero{\vt 0}

\global\long\def\bone{\mathbf{1}}

\global\long\def\boldm{\boldsymbol{m}}

\global\long\def\bff{\vt f}

\global\long\def\bx{\mathbf{x}}

\global\long\def\bc{\boldsymbol{c}}

\global\long\def\bl{\boldsymbol{l}}

\global\long\def\bu{\boldsymbol{u}}

\global\long\def\bo{\boldsymbol{o}}

\global\long\def\bk{\boldsymbol{k}}

\global\long\def\bh{\boldsymbol{h}}

\global\long\def\bs{\mathbf{s}}

\global\long\def\bz{\boldsymbol{z}}

\global\long\def\xnew{y}

\global\long\def\bxnew{\boldsymbol{y}}

\global\long\def\bX{\boldsymbol{X}}

\global\long\def\tbx{\tilde{\bx}}

\global\long\def\by{\boldsymbol{y}}

\global\long\def\bY{\boldsymbol{Y}}

\global\long\def\bZ{\boldsymbol{Z}}

\global\long\def\bU{\boldsymbol{U}}

\global\long\def\bv{\boldsymbol{v}}

\global\long\def\bn{\boldsymbol{n}}

\global\long\def\bV{\boldsymbol{V}}

\global\long\def\bK{\boldsymbol{K}}

\global\long\def\bI{\boldsymbol{I}}

\global\long\def\bw{\vt w}

\global\long\def\balpha{\gvt{\alpha}}

\global\long\def\bbeta{\gvt{\beta}}

\global\long\def\bmu{\gvt{\mu}}

\global\long\def\btheta{\boldsymbol{\theta}}

\global\long\def\blambda{\boldsymbol{\lambda}}

\global\long\def\bgamma{\boldsymbol{\gamma}}

\global\long\def\bpsi{\boldsymbol{\psi}}

\global\long\def\bphi{\boldsymbol{\phi}}

\global\long\def\bpi{\boldsymbol{\pi}}

\global\long\def\eeta{\boldsymbol{\eta}}

\global\long\def\bomega{\boldsymbol{\omega}}

\global\long\def\bepsilon{\boldsymbol{\epsilon}}

\global\long\def\btau{\boldsymbol{\tau}}

\global\long\def\bSigma{\gvt{\Sigma}}

\global\long\def\realset{\mathbb{R}}

\global\long\def\realn{\realset^{n}}

\global\long\def\integerset{\mathbb{Z}}

\global\long\def\natset{\integerset}

\global\long\def\integer{\integerset}

\global\long\def\natn{\natset^{n}}

\global\long\def\rational{\mathbb{Q}}

\global\long\def\rationaln{\rational^{n}}

\global\long\def\complexset{\mathbb{C}}

\global\long\def\comp{\complexset}

\global\long\def\compl#1{#1^{\text{c}}}

\global\long\def\and{\cap}

\global\long\def\compn{\comp^{n}}

\global\long\def\comb#1#2{\left({#1\atop #2}\right) }

\global\long\def\nchoosek#1#2{\left({#1\atop #2}\right)}

\global\long\def\param{\vt w}

\global\long\def\Param{\Theta}

\global\long\def\meanparam{\gvt{\mu}}

\global\long\def\Meanparam{\mathcal{M}}

\global\long\def\meanmap{\mathbf{m}}

\global\long\def\logpart{A}

\global\long\def\simplex{\Delta}

\global\long\def\simplexn{\simplex^{n}}

\global\long\def\dirproc{\text{DP}}

\global\long\def\ggproc{\text{GG}}

\global\long\def\DP{\text{DP}}

\global\long\def\ndp{\text{nDP}}

\global\long\def\hdp{\text{HDP}}

\global\long\def\gempdf{\text{GEM}}

\global\long\def\ei{\text{EI}}

\global\long\def\rfs{\text{RFS}}

\global\long\def\bernrfs{\text{BernoulliRFS}}

\global\long\def\poissrfs{\text{PoissonRFS}}

\global\long\def\grad{\gradient}
 \global\long\def\gradient{\nabla}

\global\long\def\partdev#1#2{\partialdev{#1}{#2}}
 \global\long\def\partialdev#1#2{\frac{\partial#1}{\partial#2}}

\global\long\def\partddev#1#2{\partialdevdev{#1}{#2}}
 \global\long\def\partialdevdev#1#2{\frac{\partial^{2}#1}{\partial#2\partial#2^{\top}}}

\global\long\def\closure{\text{cl}}

\global\long\def\cpr#1#2{\Pr\left(#1\ |\ #2\right)}

\global\long\def\var{\text{Var}}

\global\long\def\Var#1{\text{Var}\left[#1\right]}

\global\long\def\cov{\text{Cov}}

\global\long\def\Cov#1{\cov\left[ #1 \right]}

\global\long\def\COV#1#2{\underset{#2}{\cov}\left[ #1 \right]}

\global\long\def\corr{\text{Corr}}

\global\long\def\sst{\text{T}}

\global\long\def\SST{\sst}

\global\long\def\ess{\mathbb{E}}

\global\long\def\Ess#1{\ess\left[#1\right]}

\newcommandx\ESS[2][usedefault, addprefix=\global, 1=]{\underset{#2}{\ess}\left[#1\right]}

\global\long\def\fisher{\mathcal{F}}

\global\long\def\bfield{\mathcal{B}}
 \global\long\def\borel{\mathcal{B}}

\global\long\def\bernpdf{\text{Bernoulli}}

\global\long\def\betapdf{\text{Beta}}

\global\long\def\dirpdf{\text{Dir}}

\global\long\def\gammapdf{\text{Gamma}}

\global\long\def\gaussden#1#2{\text{Normal}\left(#1, #2 \right) }

\global\long\def\gauss{\mathbf{N}}

\global\long\def\gausspdf#1#2#3{\text{Normal}\left( #1 \lcabra{#2, #3}\right) }

\global\long\def\multpdf{\text{Mult}}

\global\long\def\poiss{\text{Pois}}

\global\long\def\poissonpdf{\text{Poisson}}

\global\long\def\pgpdf{\text{PG}}

\global\long\def\wshpdf{\text{Wish}}

\global\long\def\iwshpdf{\text{InvWish}}

\global\long\def\nwpdf{\text{NW}}

\global\long\def\niwpdf{\text{NIW}}

\global\long\def\studentpdf{\text{Student}}

\global\long\def\unipdf{\text{Uni}}

\global\long\def\transp#1{\transpose{#1}}
 \global\long\def\transpose#1{#1^{\mathsf{T}}}

\global\long\def\mgt{\succ}

\global\long\def\mge{\succeq}

\global\long\def\idenmat{\mathbf{I}}

\global\long\def\trace{\mathrm{tr}}

\global\long\def\argmax#1{\underset{_{#1}}{\text{argmax}} }

\global\long\def\argmin#1{\underset{_{#1}}{\text{argmin}\ } }

\global\long\def\diag{\text{diag}}

\global\long\def\norm{}

\global\long\def\spn{\text{span}}

\global\long\def\vtspace{\mathcal{V}}

\global\long\def\field{\mathcal{F}}
 \global\long\def\ffield{\mathcal{F}}

\global\long\def\inner#1#2{\left\langle #1,#2\right\rangle }
 \global\long\def\iprod#1#2{\inner{#1}{#2}}

\global\long\def\dprod#1#2{#1 \cdot#2}

\global\long\def\norm#1{\left\Vert #1\right\Vert }

\global\long\def\entro{\mathbb{H}}

\global\long\def\entropy{\mathbb{H}}

\global\long\def\Entro#1{\entro\left[#1\right]}

\global\long\def\Entropy#1{\Entro{#1}}

\global\long\def\mutinfo{\mathbb{I}}

\global\long\def\relH{\mathit{D}}

\global\long\def\reldiv#1#2{\relH\left(#1||#2\right)}

\global\long\def\KL{KL}

\global\long\def\KLdiv#1#2{\KL\left(#1\parallel#2\right)}
 \global\long\def\KLdivergence#1#2{\KL\left(#1\ \parallel\ #2\right)}

\global\long\def\crossH{\mathcal{C}}
 \global\long\def\crossentropy{\mathcal{C}}

\global\long\def\crossHxy#1#2{\crossentropy\left(#1\parallel#2\right)}

\global\long\def\breg{\text{BD}}

\global\long\def\lcabra#1{\left|#1\right.}

\global\long\def\lbra#1{\lcabra{#1}}

\global\long\def\rcabra#1{\left.#1\right|}

\global\long\def\rbra#1{\rcabra{#1}}

%% file: Introduction.tex
Bayesian optimization (BO) has received considerable attention in
tuning hyper-parameters for complex models and algorithms \cite{Brochu_2010Tutorial,Hennig_2012Entropy,Snoek_2012Practical,Shahriari_2016Taking}.
In a standard setting, BO suggests one evaluation at a time, whereas
batch BO can recommend multiple evaluations where parallel facilities
are available. The batch setting is essential to speed up the optimization.
Such scenarios appear, for instance, in optimizing computer models
where several cores are available for parallel runs. Another example
is in wet-lab experiments, wherein the need for batch experiments
is significant as the cost of testing one experiment is the same as
testing a batch.

Typically, batch Bayesian optimization considers addressing the expensive
black-box functions in which the computational cost of batch approaches
are negligible with respect to the function evaluation cost. In such
cases, the computation of batch algorithms may not influence the optimization
performance. On the contrary, if the function evaluation is cheap,
one can make use of the global optimization techniques, such as Direct
or multi-start Newton method. Thus, a computationally expensive batch
BO approach may not be the right choice when the function evaluation
is cheap.

However, there are many \emph{less expensiv}e optimization problems
where function evaluations (including evaluating time and economic
cost) are neither highly expensive nor so cheap. When optimizing
these functions, the computation of batch BO becomes sensitive and
plays an important role in the optimization performance. To highlight
such computational effects, in the experiments, we specifically consider
the less expensive functions  including machine learning models training
on medium size datasets, inexpensive physical experiment via simulators,
and acquisition function optimization (an auxiliary step in BO). The
existing batch Bayesian optimization techniques may not be well suited
for these functions. This is because their computations can be slower
than the black-box function evaluations and thus the cheap global
optimization techniques are more favorable.

In this paper, we consider batch Bayesian optimization problem for
situations where the black-box function evaluation is less expensive
in terms of evaluation time and economic cost. To make batch algorithm
efficient, we design to select a first point in the batch using a
standard BO and the remaining points in cheaper computational ways.
For this purpose, we propose a new data-driven space filling strategy,
called \emph{distance exploration }(DE)\emph{}. Our intuition for
distance exploration is based on the fact that the best location for
exploration should not be close to the existing observations. We propose
to use Sobol sequence \cite{Sobol_1967Distribution} to find such
explorative points efficiently.  Despite of being simple, our strategy
maintains the desirable property of exploration. In addition, our
model enjoys computational advantage as our model only needs to perform
a single global optimization step, as opposed to all other existing
batch Bayesian optimization approaches which either sequentially perform
$B$ global optimizations \cite{Desautels_2014Parallelizing,Kathuria_NIPS2016Batched,Contal_2013Parallel}
(where $B$ is a batch size) or solves even more complex approximation
\cite{Shah_2015Parallel,Wu_NIPS2016Parallel,Daxberger_2017Distributed}.
We validate our model using an extensive set of benchmark functions
and selected real-world applications which are less expensive. These
experiments demonstrate that our distance exploration (DE) approach
outperforms all the baselines in terms of computation while being
competitive in finding the optimal value. Our main contributions
are
\begin{itemize}
\item A first study of Bayesian optimization to tackle the problem of less
expensive function evaluation.
\item A novel view for exploration using distance and a batch BO approach
using distance exploration.
\item Validation on benchmark and real applications where evaluations are
not highly expensive.
\end{itemize}

%% file: framework_v2.tex
To address the problem of less expensive black-box function, we aim
to propose a scalable batch BO algorithm whose computation is faster
than such black-boxes. Before going to detail of the proposed method,
we briefly summarize and motivate from the existing approaches.

Most of the existing work in batch Bayesian optimization \cite{Contal_2013Parallel,Desautels_2014Parallelizing,Gonzalez_2015Batch}
try to find locations based on the shape of the posterior and the
values that one might observe at the next point being sampled. However,
BO may not reach to its asymptotic convergence because the accurate
GP posteriors can never be obtained due to (1) given limited observations
and (2) the imperfection of the GP hyper-parameter estimation, as
mentioned in \cite{Wang_2014Theoretical}. 

One way to improve Bayesian optimization is through the GP surrogate
model. To improve the GP surrogate model of the black-box function
$f$, it is intuitive to gain information about $f$ as much as possible.
By gaining more information about $f$, we can have a better fitting
of a GP surrogate. This can be done by picking the first point in
the batch to be the optimum of our standard BO, i.e., via UCB \cite{Srinivas_2010Gaussian}
or EI \cite{Mockus_1978Application} acquisition function, and choosing
the remaining points in the batch by using a cheaper strategy. Such
a cheaper strategy will help us to address the less expensive function
setting.

\subsection{The Proposed UCB-DE}

To make a batch BO algorithm efficient, we propose to select the first
element in the batch as the standard BO $\mathbf{x}_{t,1}=\arg\max_{\bx\in\mathcal{X}}\thinspace\alpha_{t}^{\textrm{UCB}}\left(\mathbf{x}\right)$.
Here, we use the GP-UCB \cite{Srinivas_2010Gaussian} which possesses
the nice convergent analysis although this can be applicable for other
acquisition functions. After obtaining the first element by UCB, we
fill in a space with a batch of $B-1$ points $\mathbf{x}_{t,b},\forall b\in\{2,...,B\}$
by using distance exploration (DE) presented in the next section.
This step aims to gather a batch of diversity points to gain information
about the black-box function $f$ in a cheap computation way. We summarize
our UCB-DE in Algorithm \ref{alg:BatchBO_Geo}. 
\begin{algorithm}
\caption{UCB-DE for Batch Bayesian Optimization.\label{alg:BatchBO_Geo}}

\begin{algor}
\item [{{*}}] Input: initial data $\mathcal{D}_{0}$, \#iter $T$, batch
size $B$, $l(\mathbf{a},\mathbf{b})=\sum_{j=1}^{d}\frac{1}{\sigma_{j}}||a_{j}-b_{j}||^{2}$,
$g(\bx,\mathcal{D})=\min_{\bx_{i}\in\mathcal{D}}l(\bx,\bx_{i})$
\end{algor}
\begin{algor}[1]
\item [{{*}}] Generate $S=\left[\bs_{1}...\bs_{M}\right]\sim\textrm{Sobol}$
sequence
\item [{for}] $t=1$ to $T$
\item [{{*}}] Obtain $\bx_{t,1}=\arg\max_{\bx\in\mathcal{X}}\thinspace\alpha_{t}^{\textrm{UCB}}\left(\bx\right)$
\item [{{*}}] Augment $\mathcal{D}_{t,1}=\mathcal{D}_{t-1}\cup\bx_{t,1}$
\item [{for}] $i=2$ to $B$
\item [{{*}}] Select $\bx_{t,i}=\arg\max_{\forall\bs\in S}g\left(\bs,\mathcal{D}_{t,i-1}\right)$
\item [{{*}}] Augment $\mathcal{D}_{t,i}=\mathcal{D}_{t,i-1}\cup\bx_{t,i}$
\item [{endfor}]~
\item [{{*}}] Evaluate in parallel $y_{t,b}=f\left(\bx_{t,b}\right),\forall b\le B$ 
\item [{{*}}] Augment $\mathcal{D}_{t}=\mathcal{D}_{t-1}\cup\left(\bx_{t,b},y_{t,b}\right)_{b=1}^{B}$
\item [{endfor}]~
\end{algor}
\begin{algor}
\item [{{*}}] Output: a recommendation $\bx^{*}=\arg\max_{\bx\in\mathcal{X}}\mu(\bx\mid\mathcal{D}_{T})$
\end{algor}
\end{algorithm}

Our approach spends $B-1$ points in a batch to gain information about
the black-box function $f$. We guide it to select the points which
are as far as possible from the existing observations to learn about
$f$. This way will help to estimate better the GP model (a surrogate
model of $f$) which later informs to select a next point. Since the
first point in a batch is using a standard BO, our model still performs
exploitation to find the optimum, one of the key factor in decision
making. In other words, our approach uses $B-1$ points for exploration
and a point for exploitation in each iteration.

Our approach requires only one run of global optimization, opposed
to all other existing batch BO approaches which either sequentially
perform $B$ global optimizations \cite{Desautels_2014Parallelizing,Kathuria_NIPS2016Batched,Contal_2013Parallel}
or solve even more complex approximation \cite{Shah_2015Parallel,Wu_NIPS2016Parallel,Nguyen_ICDM2016Budgeted,Daxberger_2017Distributed,wang2018batched}.

\subsection{Distance Exploration (DE)}

We present a simple, but effective, strategy for filling the $B-1$
remaining points in a batch, called distance exploration. We first
summarize that the traditional space filling strategies can not take
into account the existing observations to make the best space-filling.
Then, we describe the proposed distance exploration and discuss some
useful properties.

\subsubsection{Data-driven space filling strategies. }

\begin{figure}
\begin{centering}
\includegraphics[width=0.95\columnwidth]{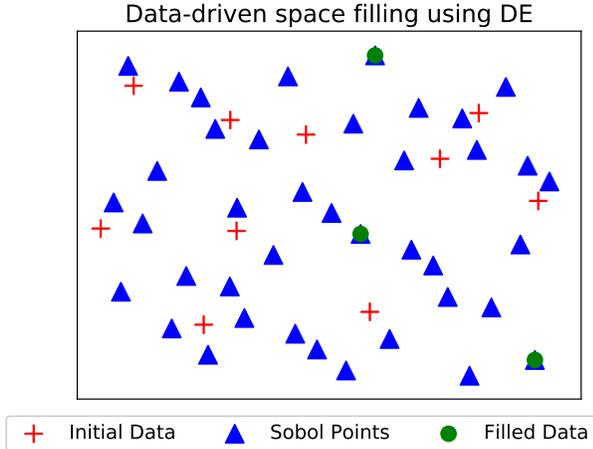}
\par\end{centering}
\caption{Examples of DE to fill a space with $3$ extra points given $10$
initial observed locations (\textcolor{red}{$+$}) . We first use
a  Sobol sequence \textcolor{blue}{$\Delta$} of $40$ points to
approximately find the good location which is far from the existing
observations. Then, we pick the best point from this Sobol sequence
and add this point into the observation set. We repeat the process
until a batch is filled. This process is done without requiring a
global optimization.\label{fig:Examles-of-space-filling-1}}
\end{figure}

There are existing well known strategies for space filling, such as
Latin hypercube design \cite{Mckay_1979comparison} and Sobol sequence
\cite{Sobol_1967Distribution}. However, most of such strategies are
not designed for data-driven setting, or under the presence of observations.
Example of the Sobol sequence is in Fig. \ref{fig:Examles-of-space-filling-1}.
We can see that the points in the vanilla Sobol sequence are not yet
optimal for exploration as they may locate near the existing observations.
This is because the vanilla Sobol sequence is unable to take into
account the existing observations to make a better design. In our
setting of batch BO, we aim to fill a space such that we do not want
to assign points close to the existing observations. The conventional
space filling approaches \cite{Sobol_1967Distribution} are restricted
for this scenario, to our knowledge.

\subsubsection{Distance exploration (DE) for data-driven space filling.}

We propose a space filling approach given the existence of the observed
locations. Intuitively, we aim to fill in a space with high uncertainty
points which should not be close to already observed locations. If
a considered data point stays closer to an observed location, the
less uncertainty it gets, and vice versa. Thus, we can employ the
distance from an arbitrary location to the observed locations as a
guide for filling a space. That is, we sequentially fill a space with
the location $\mathbf{x}_{t}$ s.t. the distance from $\mathbf{x}_{t}$
to its nearest observation $\mathbf{x}_{i}\in\mathcal{D}_{t}$ is
maximized, i.e. $\mathbf{x}_{t}=\arg\max_{\mathbf{x}\in\mathcal{X}}\thinspace||\mathbf{x}-[\mathbf{x}]||^{2}$
where we denote a nearest observation to $\mathbf{x}$ as $[\mathbf{x}]=\arg\min_{\bx_{i}\in\mathcal{D}_{t}}||\mathbf{x}-\mathbf{x}_{i}||^{2}$
and $\mathcal{D}_{t}$ is the observation set at iteration $t$. By
using distance for data-driven space filling, our DE offers the alternative
exploration to a Gaussian process predictive variance while DE gets
rid of cubic operation of these methods. Formally, we seek for the
point which is far from the existing observations, i.e.,
\begin{align}
\mathbf{x}_{t} & =\argmax{\mathbf{x}\in\mathcal{X}}\thinspace[\argmin{\mathbf{x}_{i}\in\mathcal{D}_{t}}l(\mathbf{x},\mathbf{x}_{i})].\label{eq:DE_maximization_min}
\end{align}
where the $\arg\min$ is computed from a finite set of $\mathcal{D}_{t}$
and $\arg\max$ can be computed using either a global optimization
on a continuous domain $\mathcal{X}$ or a Sobol approximation in
the below section.

\subsection{Finding The Farthest Points Efficiently}

One can optimize the Eq. (\ref{eq:DE_maximization_min}) directly
to find the farthest point, $\mathbf{x}\in\mathcal{X}$, from the
existing observation set $\mathcal{D}_{t}$ by running a global optimization.
However, running $B-1$ global optimizations will be expensive, especially
for high dimensional functions, and finding the exact locations may
not be necessary since we will not directly use such points for final
recommendation. Instead, these points are used to gain information
for estimating a better GP surrogate model.

Therefore, we propose an efficient algorithm to find the farthest
points without using global optimization. Our idea is as the following.
Initially, we generate a Sobol sequence \cite{Sobol_1967Distribution}
including $M$ points, $S=[\bs_{1},...,\bs_{M}],\bs_{m}\in\mathcal{R}^{d}$.
This step is required to compute once in advance. Then, at each iteration
$t$, we evaluate the ``explorative score'' at those Sobol points
using a function $\min_{\bx_{i}\in\mathcal{D}_{t}}l(\bs_{m},\mathbf{x}_{i})$.
Based on such scores, we can approximately compute the Eq. (\ref{eq:DE_maximization_min})
to choose the farthest point as
\begin{align}
\mathbf{x}_{t} & =\argmax{\forall\bs_{m}\in S}[\argmin{\bx_{i}\in\mathcal{D}_{t}}l(\mathbf{x},\mathbf{x}_{i})].\label{eq:DE_Sobol-1}
\end{align}

Next, we add this found point into the observation set (in a greedy
manner) $D_{t,i}=D_{t,i-1}\cup\mathbf{x}_{t}$ and sequentially repeat
this process to fill a remaining $B-1$ points in a batch while the
first point in a batch will be chosen by the standard GP-UCB.

An example of distance exploration for space filling is illustrated
in Fig. \ref{fig:Examles-of-space-filling-1}. Given the $10$ initial
observed locations (\textcolor{red}{$+$}) $\mathcal{D}_{t}$, we
generate $40$ Sobol sequences (\textcolor{blue}{$\Delta$}) then
we fill a space with $3$ additional points by choosing the best points
from the Sobol sequence.

\subsection{Computational Complexity \label{subsec:Computational-Complexity}}

Let $T$ denote \#iteration, $B$ \#batch size, $d$ \#dimension,
$N=TB$ \#observation. In BO, global optimization, which repeatedly
evaluates the acquisition function, is used to select a point. Let
the number of evaluations in global optimization be $C_{d}$ as a
function of the dimension $d$. Both BUCB \cite{Desautels_2014Parallelizing}
and our approach are similar in computing the first point in a batch
with the complexity of $\mathcal{O}(C_{d}TN^{2})$ where the Cholesky
decomposition is used for matrix inversion. 

The difference in computation is from selecting $B-1$ points. In
BUCB, , the total cost for selecting $B-1$ elements in $T$ iteration
is $\mathcal{O}\left[T(B-1)(N^{2}+C_{d}N^{2})\right]$. By noting
that $N=TB$, we have $\mathcal{O}(C_{d}N^{3})$ and thus the total
complexity of BUCB (also applicable for CL, UCB-PE and DPP) is $\mathcal{O}(C_{d}\left[TN^{2}+N^{3}\right])$.

In our UCB-DE, the computation of Sobol sequence is pre-computed once.
The cost for finding a nearest point in $\mathcal{D}_{t}$ is also
cheap $\mathcal{O}(N)$.  Thus, our UCB-DE is $\mathcal{O}(C_{d}TN^{2})$,
much cheaper than BUCB.

\subsection{Evaluation of Performance \label{subsec:Evaluation-of-Performance}}

In our setting, the points selected by the distance exploration are
different from the posterior maximum and does not represent the best
guess for the global minimizer at each step. On the other hand, when
using the batch strategy based on the values observed, such as GP-BUCB
\cite{Desautels_2014Parallelizing} and LP \cite{Gonzalez_2015Batch},
these points do still provide good results. However, when using distance
exploration, the chosen points do not necessary contain high function
values. We therefore propose a greedy evaluation at the posterior
minimum as the final step of optimization. That is, after $t$ iterations,
we take the recommendation point as $\tilde{\mathbf{x}}_{t}=\arg\max_{\mathbf{x}\in\mathcal{X}}\mu(\mathbf{x}\mid\mathcal{D}_{t})$
and use $f(\tilde{\mathbf{x}}_{t})$ for comparison. This strategy
is also popularly used for evaluating information-theoretic BO approaches
\cite{Hennig_2012Entropy,Hernandez_2014Predictive,ru2018fast}. 
\begin{figure*}
\begin{centering}
\includegraphics[width=0.66\columnwidth]{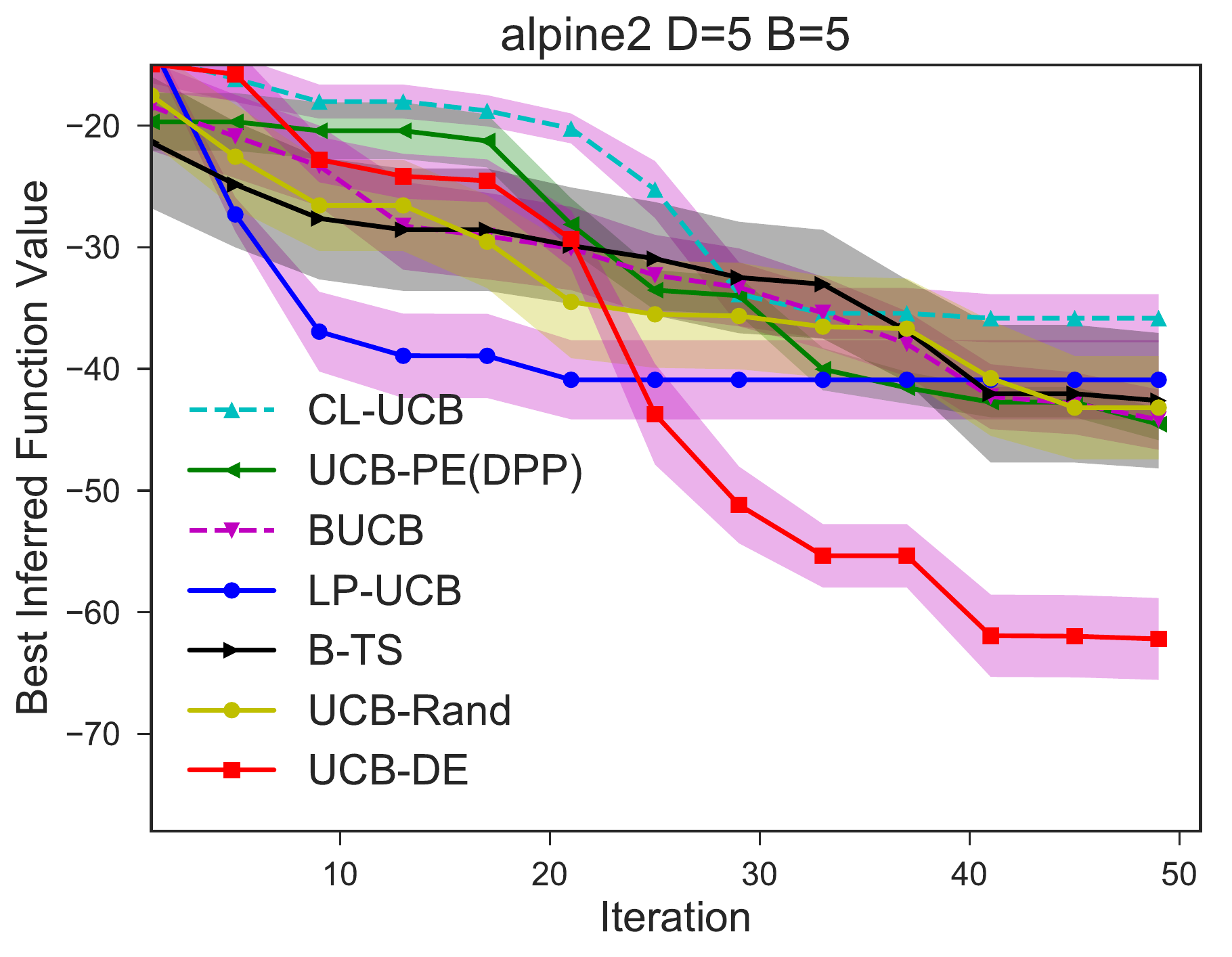}\includegraphics[width=0.67\columnwidth]{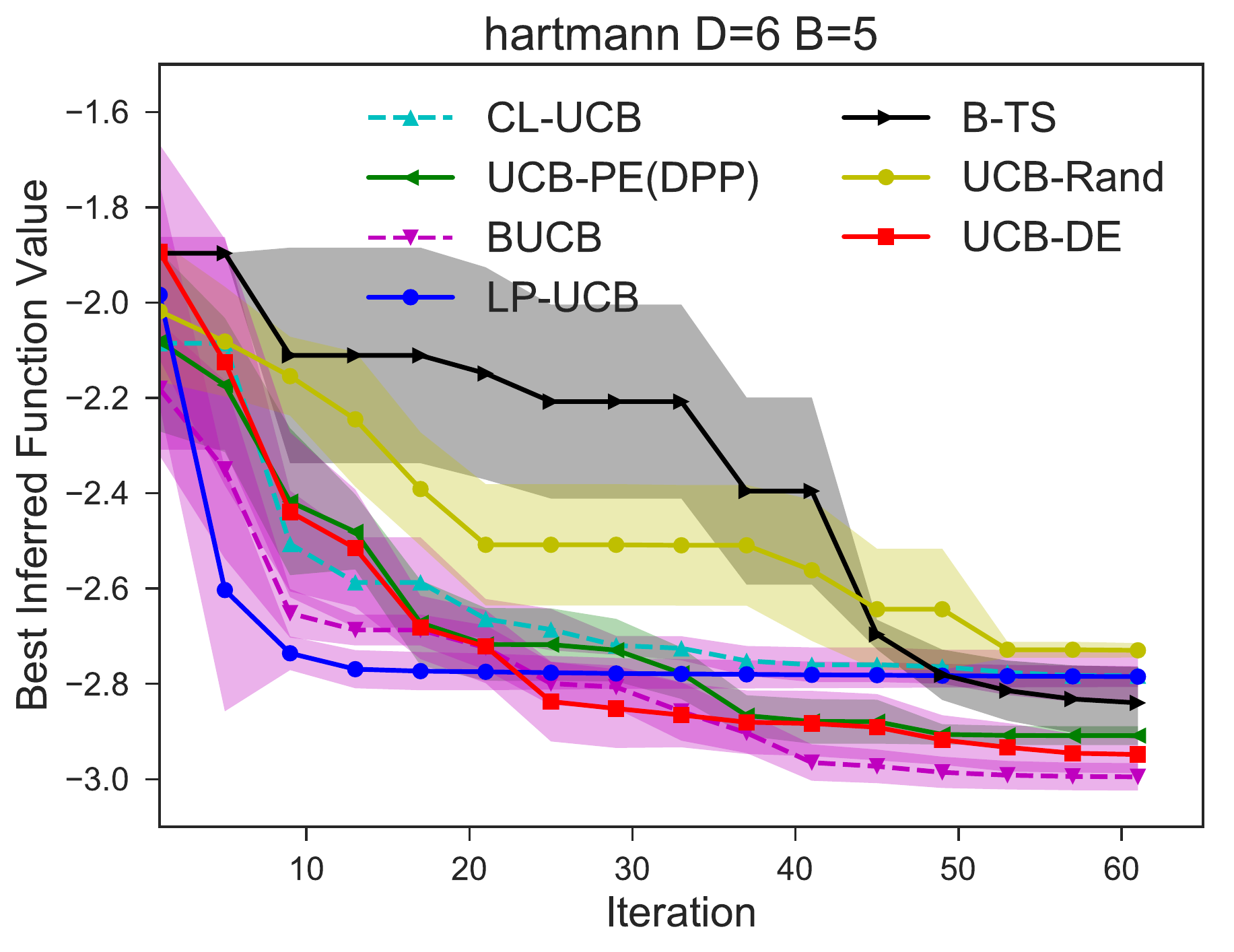}\includegraphics[width=0.66\columnwidth]{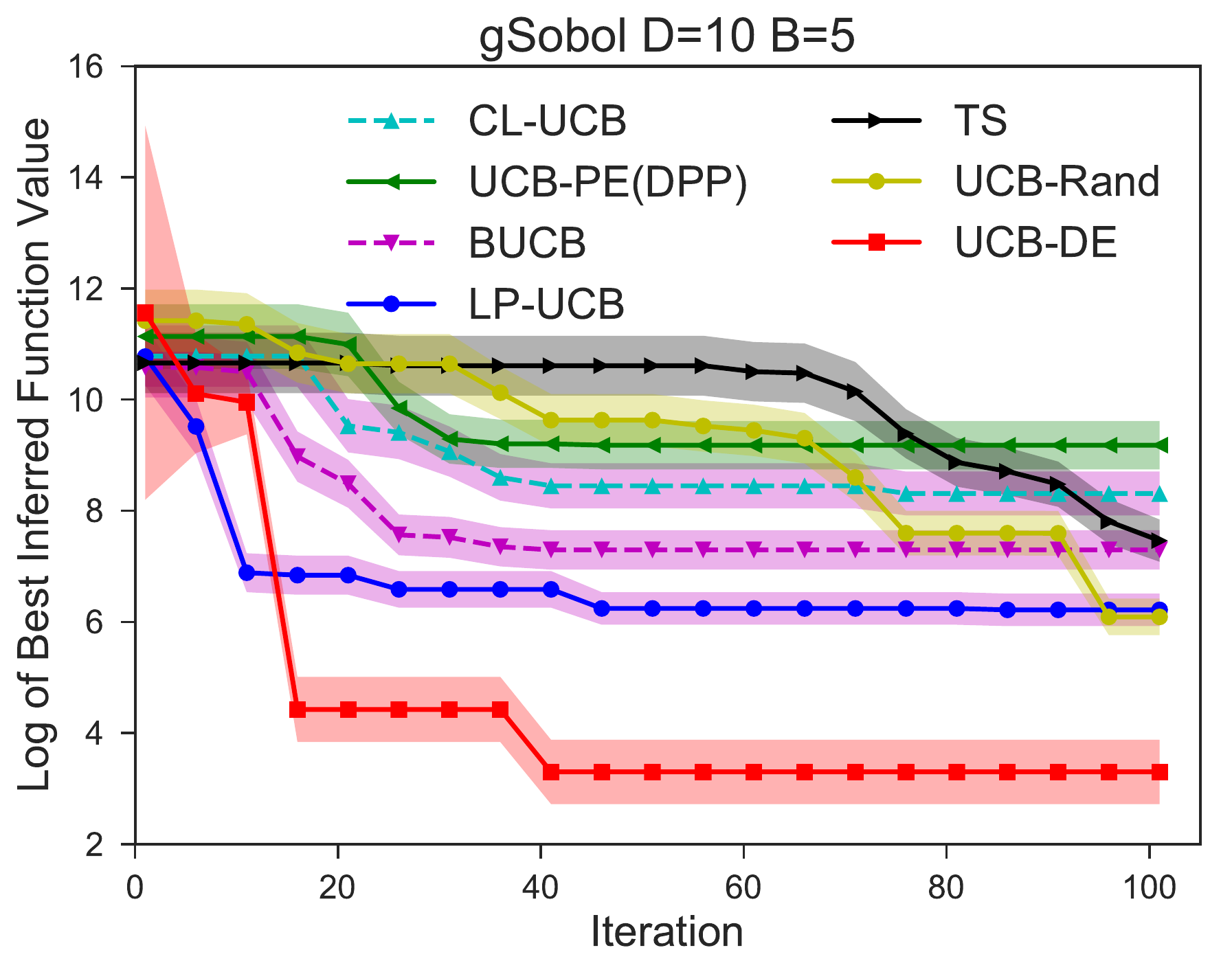}
\par\end{centering}
\begin{centering}
\includegraphics[width=0.66\columnwidth]{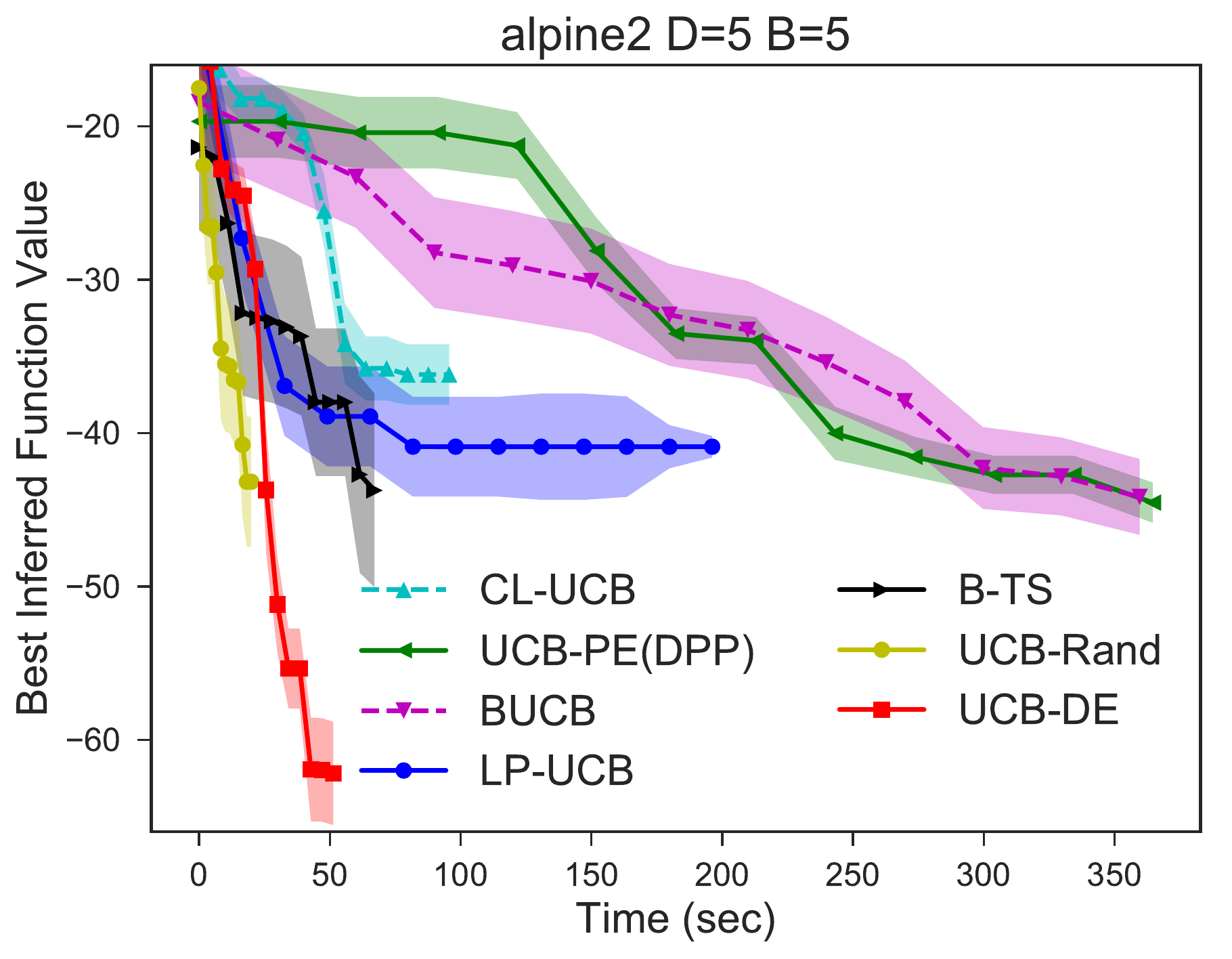}\includegraphics[width=0.66\columnwidth]{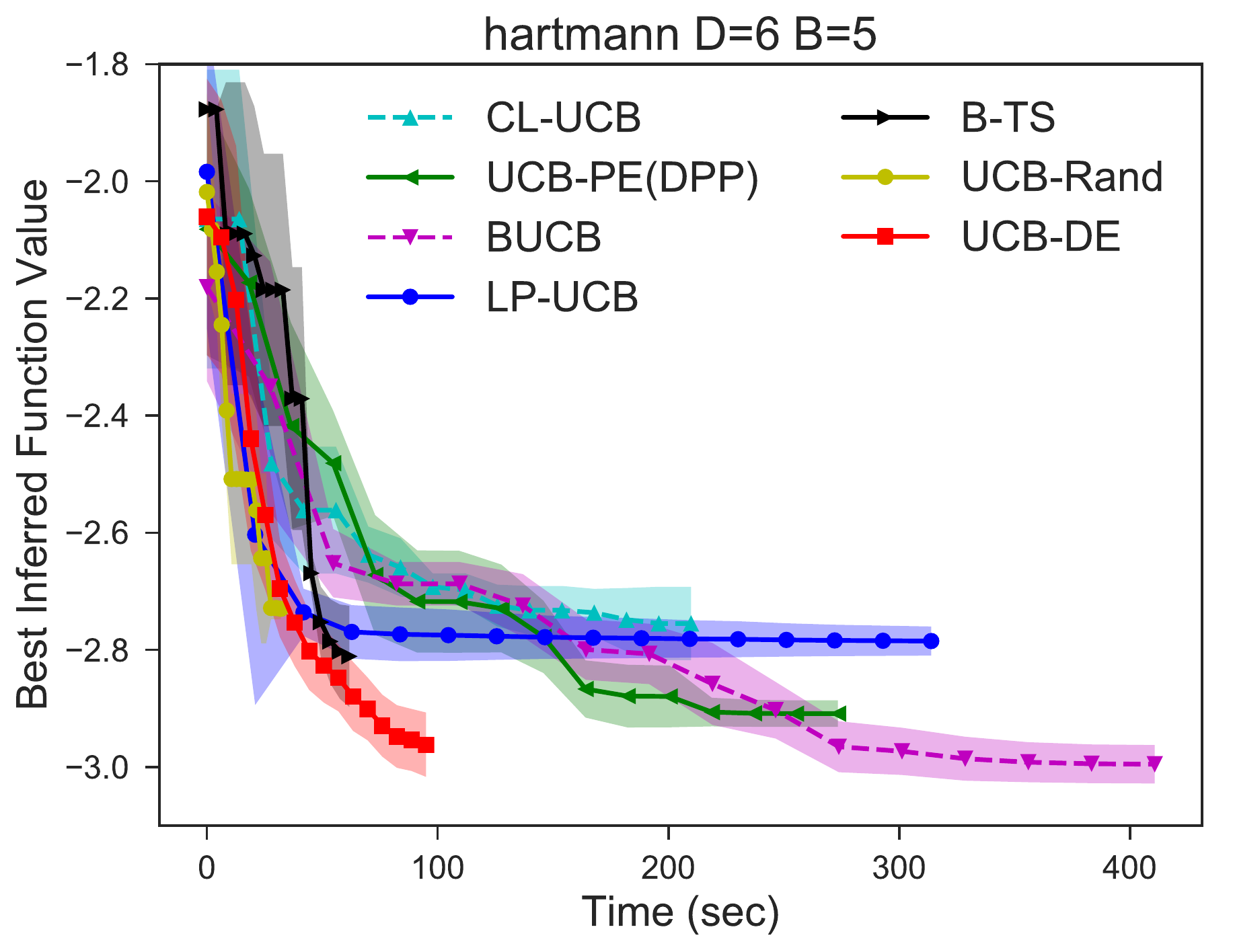}\includegraphics[width=0.66\columnwidth]{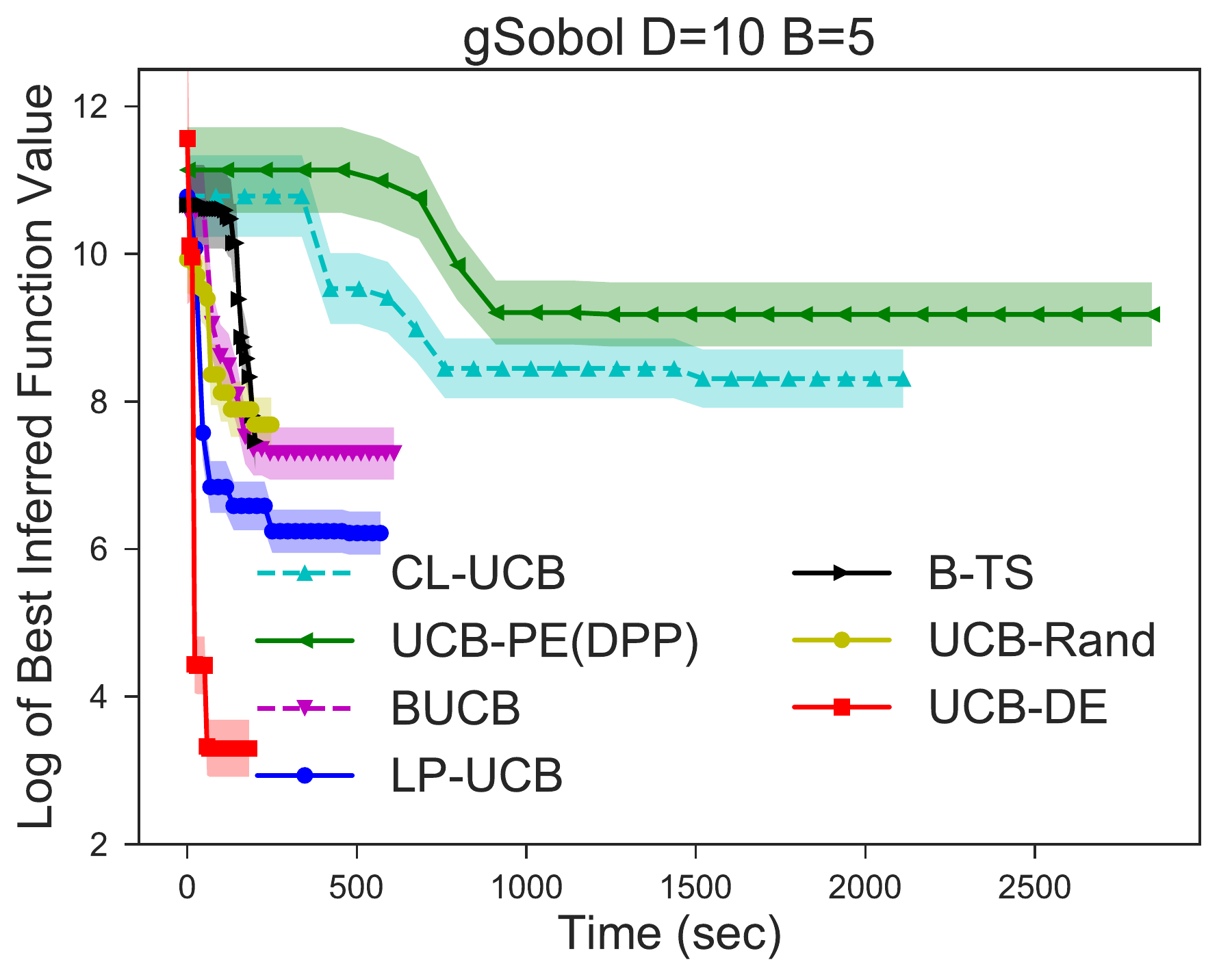}
\par\end{centering}
\caption{Top: comparison on iteration axis. Bottom: comparison on time axis
which is suitable for a setting of less expensive function. Our approach
achieves superior performance within limited time. Best viewed in
color. \label{fig:Performance-comparison-w.r.t_time-1}}
\end{figure*}

%% file: Experiments_v1.tex
We first demonstrate the efficiency of our UCB-DE against the baselines
on benchmark functions. Next, we highlight our model on optimizing
the less expensive functions. Finally, we analyze the computational
complexity. 

\vspace{-0pt}

\paragraph*{Experimental setting. \label{par:Experimental-setting-1}}

Given dimension $d$, the optimization is run with an evaluation budget
of $T=10d$ and the initial $3d$ points. The input is scaled $\mathbf{x}\sim\left[0,1\right]^{d}$
and the output is standardized $y\sim\mathcal{N}\left(0,1\right)$
for robustness. Although our DE is applicable for anisotropic case
of the distance, we use the isotropic Euclidean distance for simplicity
since our main goal is to demonstrate the scalability for optimizing
less expensive function. For other BO approaches, we use the isotropic
squared exponential (SE) kernel $k\left(\mathbf{x},\mathbf{x'}\right)=\exp\left(-l(\mathbf{x},\mathbf{x'})|^{2}\right)$
where $l(\mathbf{a},\mathbf{b})=\frac{1}{\sigma_{l}}||\mathbf{a}-\mathbf{b}||^{2}$
and $\sigma_{l}$ is chosen by maximizing the GP marginal likelihood
\cite{Rasmussen_2006gaussian}. We repeat the experiments $20$ times
and report the mean and standard error. All implementations are
in Python on a Windows machine Core i7 Ram 24GB. For the experimental
evaluation, after $t$ iterations, we take the recommendation point
as the maximum of the GP posterior, i.e., $\tilde{\mathbf{x}}_{t}=\arg\max_{\mathbf{x}\in\mathcal{X}}\mu(\mathbf{x}\mid\mathcal{D}_{t})$
and use $f(\tilde{\mathbf{x}}_{t})$ for comparison. We note that
other batch approaches, such as BUCB, using either $f(\tilde{\mathbf{x}}_{t})$
or $\max_{\forall i\le t}f(\mathbf{x}_{t})$ will result in similar
performance. We use a Sobol sequence with $M=10\times T\times B$
points. 
\begin{figure}
\begin{centering}
\includegraphics[width=1\columnwidth]{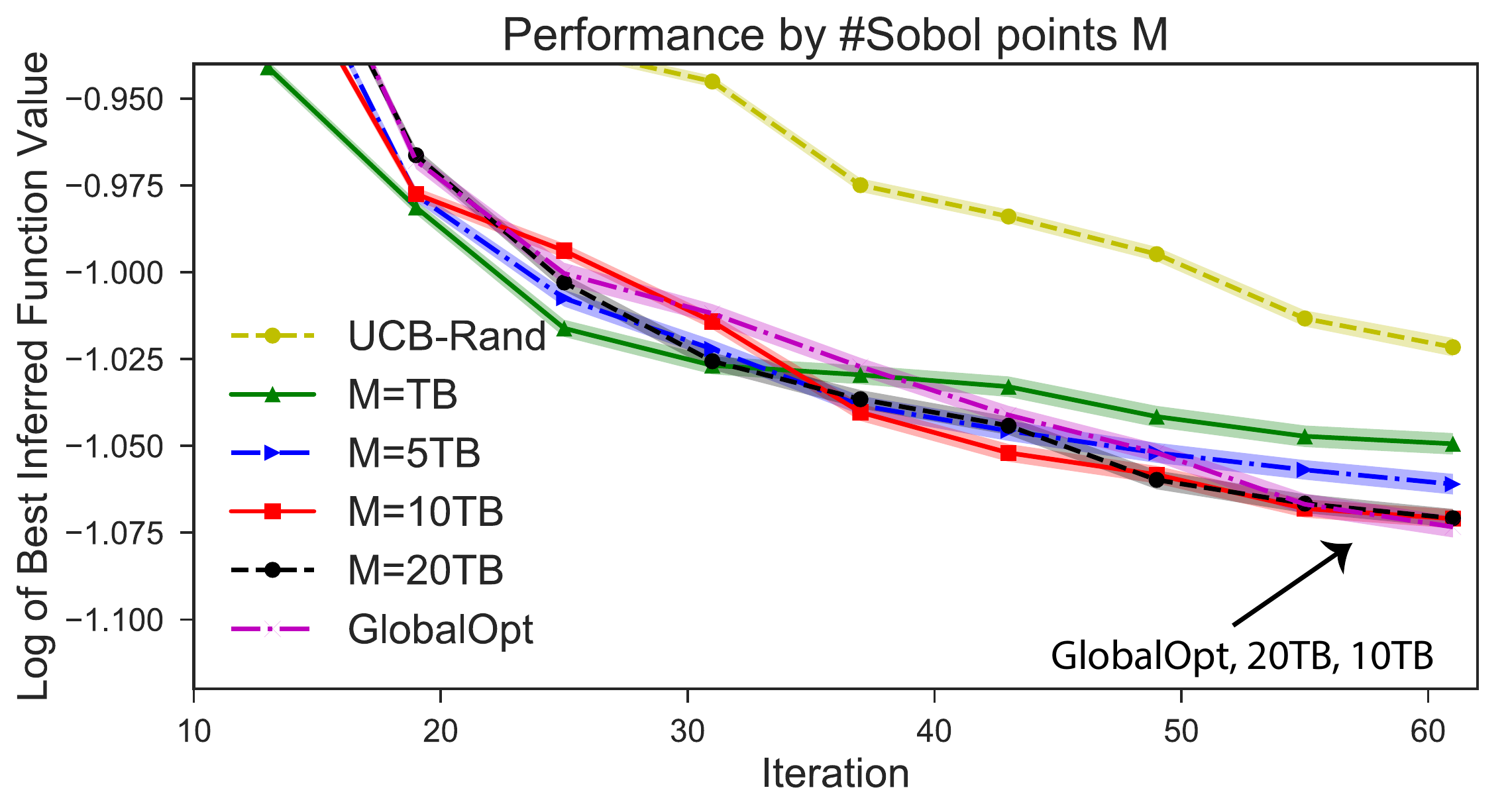}
\par\end{centering}
\caption{Comparison with different choices of $M$. Our approach UCB-DE using
Sobol sequence $M\ge10TB$ achieves comparable performance to using
continuous global optimization.\label{fig:UCB-DE_withM}}
\end{figure}

\paragraph{Baselines.}

We use  common batch BO baselines and global optimization methods
for comparison. These baselines include\emph{ Constant liar (CL)
\cite{Ginsbourger_2010Kriging} }and\emph{ local penalization (LP)
}\cite{Gonzalez_2015Batch} from GPyOpt toolbox\footnote{https://github.com/SheffieldML/GPyOpt}.\emph{
GP-BUCB} \cite{Desautels_2014Parallelizing},\emph{ UCB pure exploration
(PE)} \cite{Contal_2013Parallel} and \emph{determinantal point process
(DPP)} \cite{Kathuria_NIPS2016Batched}. The UCB-PE is shown as equivalent
to the greedy DPP. Thus, in the experiment, we denote UCB-PE (DPP)
for comparing these two baselines.\emph{ UCB-Rand: }select a first
point in a batch by GP-UCB and the remaining points by random. \emph{Batch
Thompson sampling (B-TS)}:  although the batch TS approach \cite{Hernandez_2017Parallel,Kandasamy_2018Parallelised}
is presented for distributed setting, we restrict B-TS non-distributed
in this paper.\emph{ Random search, L-BFGS-B} (with multi-start) in
Scipy and \emph{Direct} in nlopt package \cite{johnson2014nlopt}.

\subsection{Model analysis on number of Sobol points $M$\label{subsec:Model-analysis-onM}}

We analyze our model performance by varying the number of Sobol points
in Fig. \ref{fig:UCB-DE_withM}. Using Hartmann $d=6$ function, we
can see that increasing $M$ can likely improve the performance which
is converged after $M\ge10TB$. Although we fix the number of samples
$M=10TB$ in the experiments, we would suggest increasing this value
if we have generous computational budget.

In addition, we empirically show that our approach using Sobol sequence
in Eq. (\ref{eq:DE_Sobol-1}) with sufficiently number of points $M$
(e.g., $10TB$ or $20TB$) achieves comparable performance to the
case of using continuous global optimization, as in Eq. (\ref{eq:DE_maximization_min}).
The first reason is by the imperfectness of the global optimization.
The second reason is that we may not need to find the exact farthest
points for the purpose of exploration. 

The UCB-Rand is inferior to UCB-DE irrespective of the choices $M$.
This is because of the property of Sobol sequence \cite{Sobol_1967Distribution}
makes the points diversed while there is no such guarantee for random
points.

\subsection{Comparison on benchmark functions}

\begin{figure}
\begin{centering}
\includegraphics[width=0.94\columnwidth]{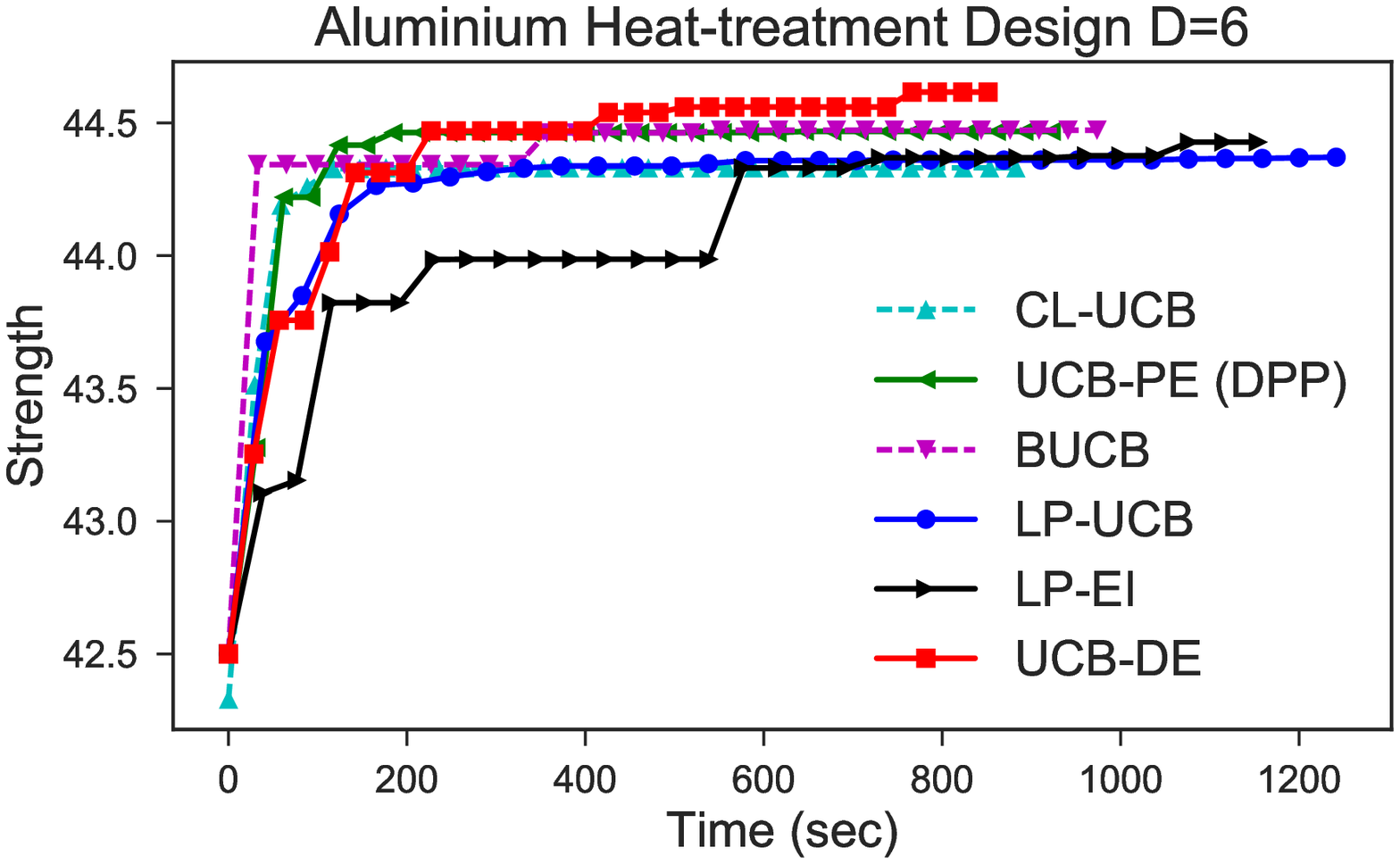}
\par\end{centering}
\begin{centering}
\includegraphics[width=0.94\columnwidth]{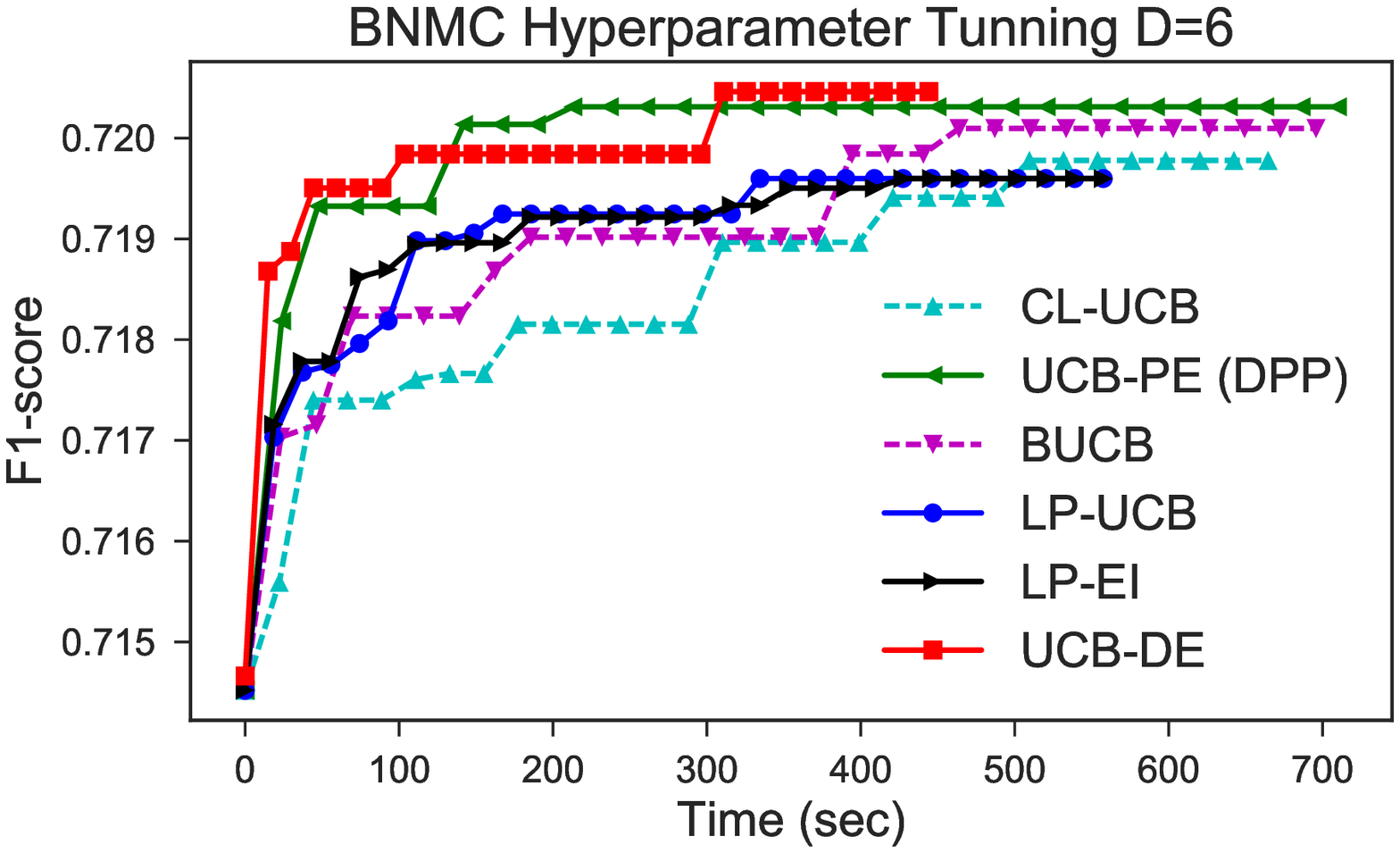}
\par\end{centering}
\begin{centering}
\includegraphics[width=0.94\columnwidth]{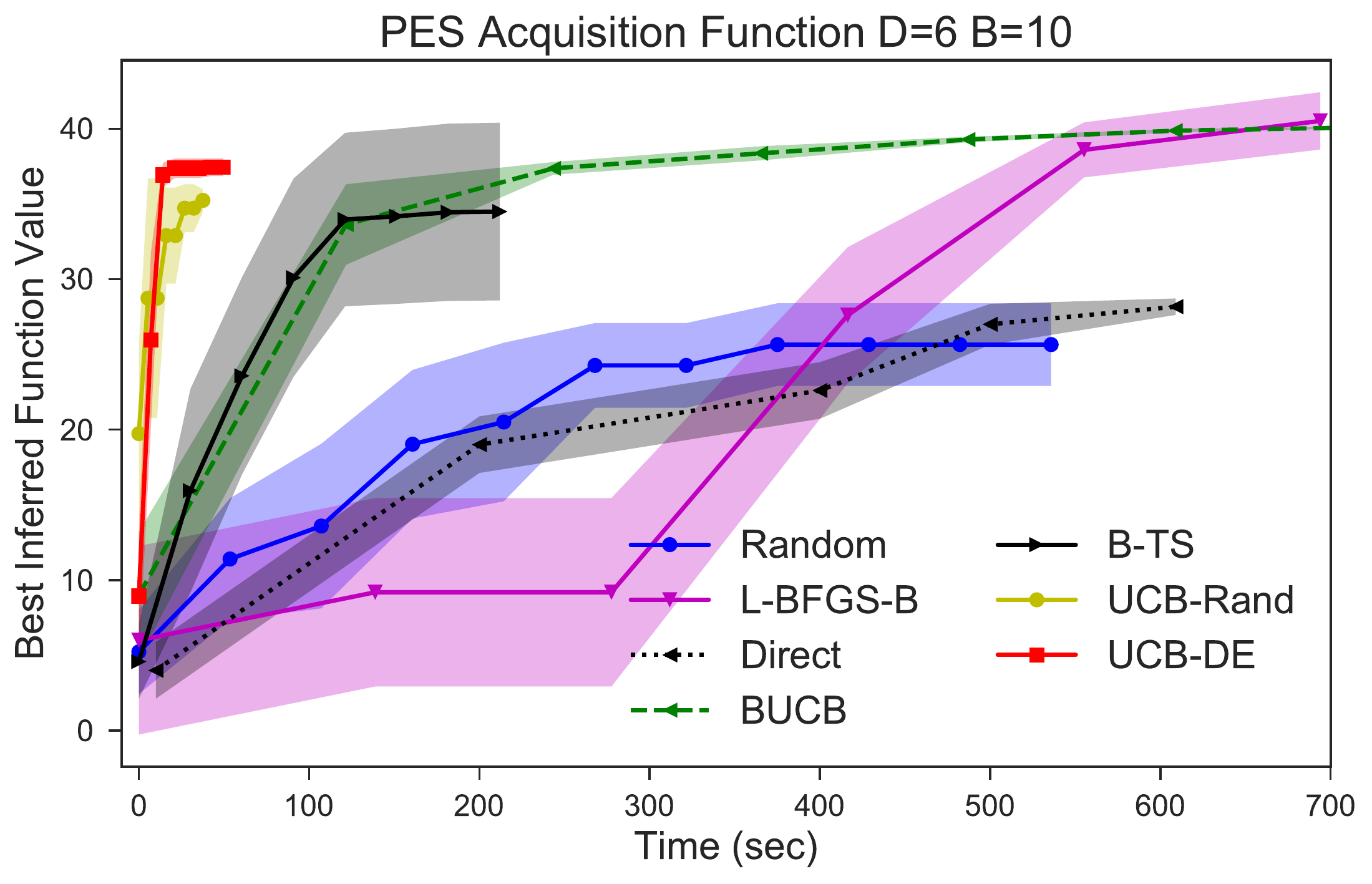}
\par\end{centering}
\begin{centering}
\includegraphics[width=0.94\columnwidth]{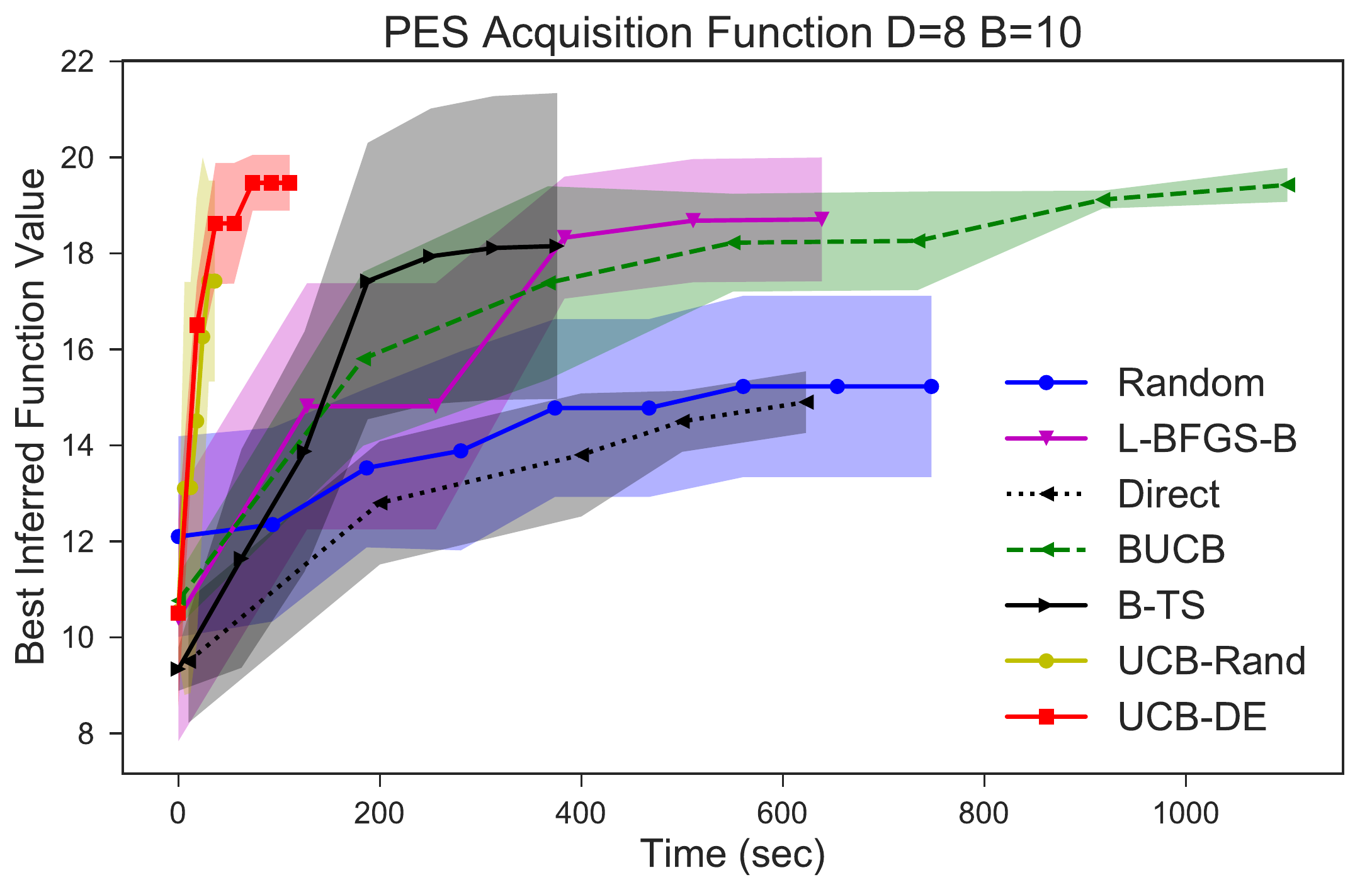}\caption{Optimizing less expensive black-box functions in a time axis (including
the time for BO and for evaluating experiments in parallel). UCB-DE
outperforms the baselines within the shortest time. When function
evaluations are a bit more expensive (top row), the BO approaches
are clearly better than the global optimization.\label{fig:Batch-BO-comparison_real_exp-1}}
\par\end{centering}
\vspace{-1pt}
\end{figure}
We compare our UCB-DE with batch BO baselines on benchmark functions
in Fig. \ref{fig:Performance-comparison-w.r.t_time-1} under the iteration
axis (top) and time axis (bottom). All of the BO methods perform
generally well and competitive because they are similar in selecting
the first element in a batch (except B-TS). The key difference is
from selecting the remaining $B-1$ points. Our proposed UCB-DE 
performs better than the baselines. This is because DE selects points
far away from the existing observations to learn better the GP surrogate
model. We improve the optimization by improving our surrogate model.
In addition, selecting points by distance exploration is beneficial
against the GP predictive variance in such a way that GP variance
is crucially sensitive to the choice of kernel length-scale parameter
which may not be estimated accurately given limited observations.
As a result, our UCB-DE gains favorable performance against the baselines.
In particular, for alpine2 and gSobol functions which require more
exploration, our UCB-DE significantly outperforms all competitors
by a wide margin.

Another useful property of our UCB-DE is the computational advantage.
Using a time axis (Fig. \ref{fig:Performance-comparison-w.r.t_time-1}
bottom), we highlight that our UCB-DE outperforms the others within
a short period. UCB-DE is better than UCB-Rand because the DE can
gather information optimally than random which can select points near
the existing observations.

\subsection{Batch BO for less expensive experiments \label{subsec:Machine-Learning-Hyperparameter}}

We demonstrate the effectiveness of our UCB-DE on real-world applications
where the function evaluations are less expensive. Specifically, we
pick the experiments such that each experiment evaluation is in a
range of $1-10$ times the CPU computation of the batch BO algorithm.
We select three experiments with the decreasing expensiveness order
as (1) physical heat-treatment design, (2) machine learning model,
and (3) optimizing the PES acquisition functions. We use a batch
size $B=10$ and the number of iteration is $T=10d$ for the batch
approaches whereas the global optimization solvers can take more evaluations
given the same time budget.

\paragraph*{Heat-treatment design for Aluminum.}

We consider the alloy hardening process of Aluminum-scandium \cite{Kampmann_1983Kinetics}
consisting of three stages each of which includes a choice of time
and temperature. We aim to maximize the strength for alloys by designing
the appropriate times and temperatures. Thus, we have 6 experimental
design choices to tune.  We collaborate with the metallurgists to
evaluate the strength using the KWN Matlab simulator. Each evaluation
takes about $30$ secs.

\paragraph*{Tuning cheap machine learning model.}

We optimize the hyper-parameters for the multi-label classification
algorithm of BNMC \cite{Nguyen_ACML2016Bayesian} on Scene dataset
using the available code. There are 6 hyper-parameters to tune so
that the F1-score is maximized. Each evaluation takes $20$ secs.

\paragraph*{PES acquisition function.}

Optimizing the acquisition function to select a next evaluation is
a common strategy in BO. Gradient-based optimizer is ideal for this
step when the acquisition function has the derivative information
available (e.g., EI and UCB). However, it is not the case for information
theoretic acquisition functions, such as Predictive entropy search
\cite{Hernandez_2014Predictive} because of multiple assumptions
and approximation by expectation propagation which does not guarantee
to converge and often leads to numerical instabilities. Formally,
we consider $\alpha^{\textrm{PES}}$ a less expensive black-box function
and use batch BO to optimize $\mathbf{x}_{t}=\arg\max\alpha^{\textrm{PES}}(\mathbf{x})$.
Each evaluation of $\alpha^{\textrm{PES}}(\mathbf{x})$ takes $2$
secs although finding $\arg\max\alpha^{\textrm{PES}}(\mathbf{x})$,
requiring many evaluations, will take $800-3000$ secs.

\paragraph*{Analyzing the results.}

We report the results in Fig. \ref{fig:Batch-BO-comparison_real_exp-1}.
It can be seen that the performance of B-TS and Random approaches
are in high uncertainty than the others. Although the Random, L-BFGS-B
(with multi-start) and Direct can take more evaluations, they generally
perform worse than the BO approaches. This is because BO approaches,
with theoretical guarantee, can balance exploration and exploitation
to reach to the better location using less evaluations. 

When function evaluations are a bit more expensive (i.e. heat-treatment
and BNMC in the Top Fig. \ref{fig:Batch-BO-comparison_real_exp-1}),
the BO approaches are clearly better than all global optimization
approaches. On the other hand, for cheaper evaluation functions of
PES, the L-BFGS-B (with multi-start) will perform well (see Bottom
Fig. \ref{fig:Batch-BO-comparison_real_exp-1}). 

The experiments show that our UCB-DE overall achieves  the best performance
within the shortest time (see Fig. \ref{fig:Batch-BO-comparison_real_exp-1}).
Although BUCB also gains good optimization values (similar to ours)
at the end, it takes BUCB $3-6$ times slower than our UCB-DE to reach
there. We remark that a random search (Random in Fig. \ref{fig:Batch-BO-comparison_real_exp-1})
results in poor performance especially in high dimension whereas a
random search with UCB (or UCB-Rand) performs surprisingly well. Specifically,
we demonstrate that using our UCB-DE can speed up the optimization
cost for PES acquisition function \cite{Hernandez_2014Predictive}
from $800$ secs in $d=6$  to $100$ secs for the same optimal value.
\begin{figure}
\begin{centering}
\includegraphics[width=1\columnwidth]{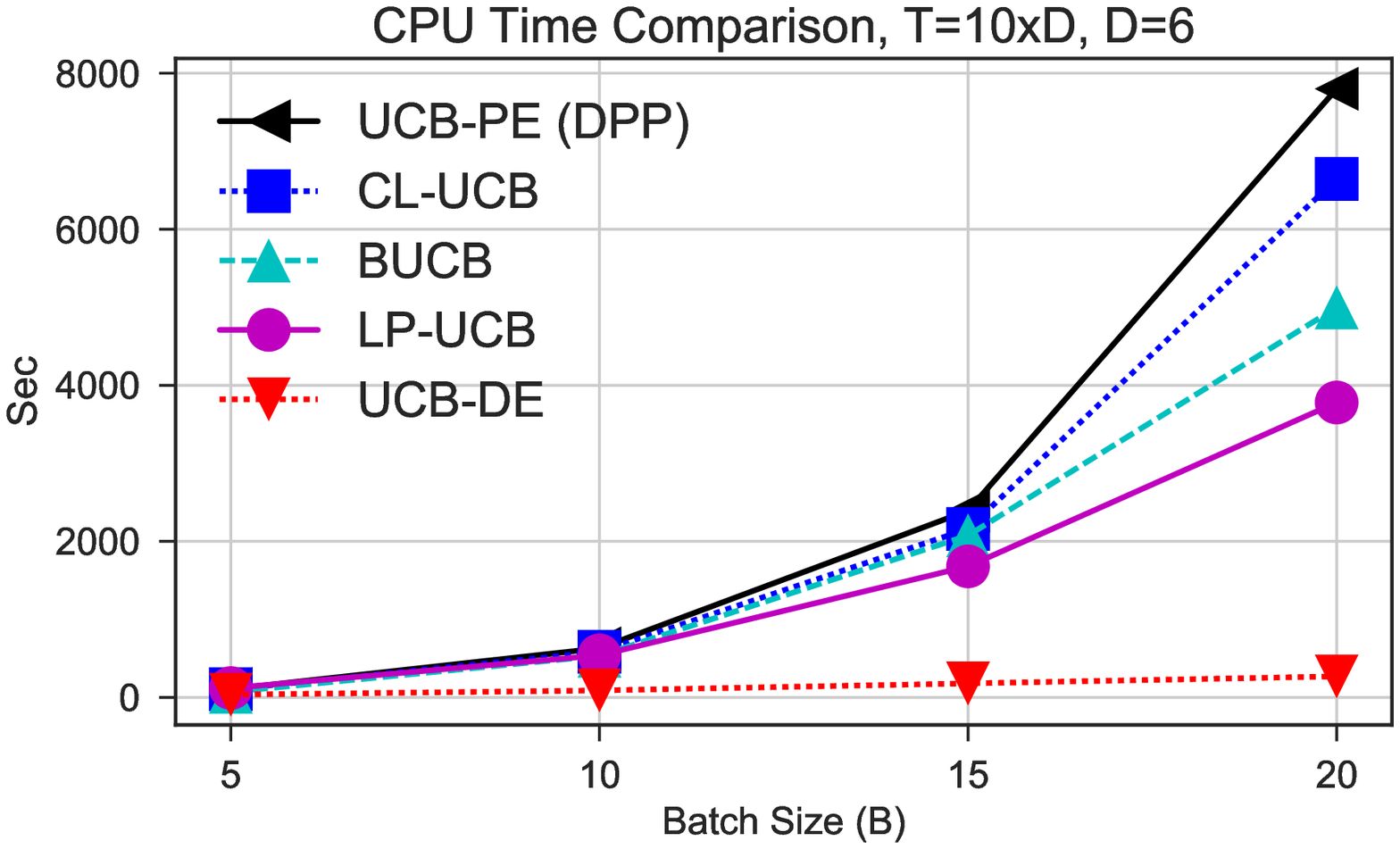}
\par\end{centering}
\begin{centering}
\includegraphics[width=1\columnwidth]{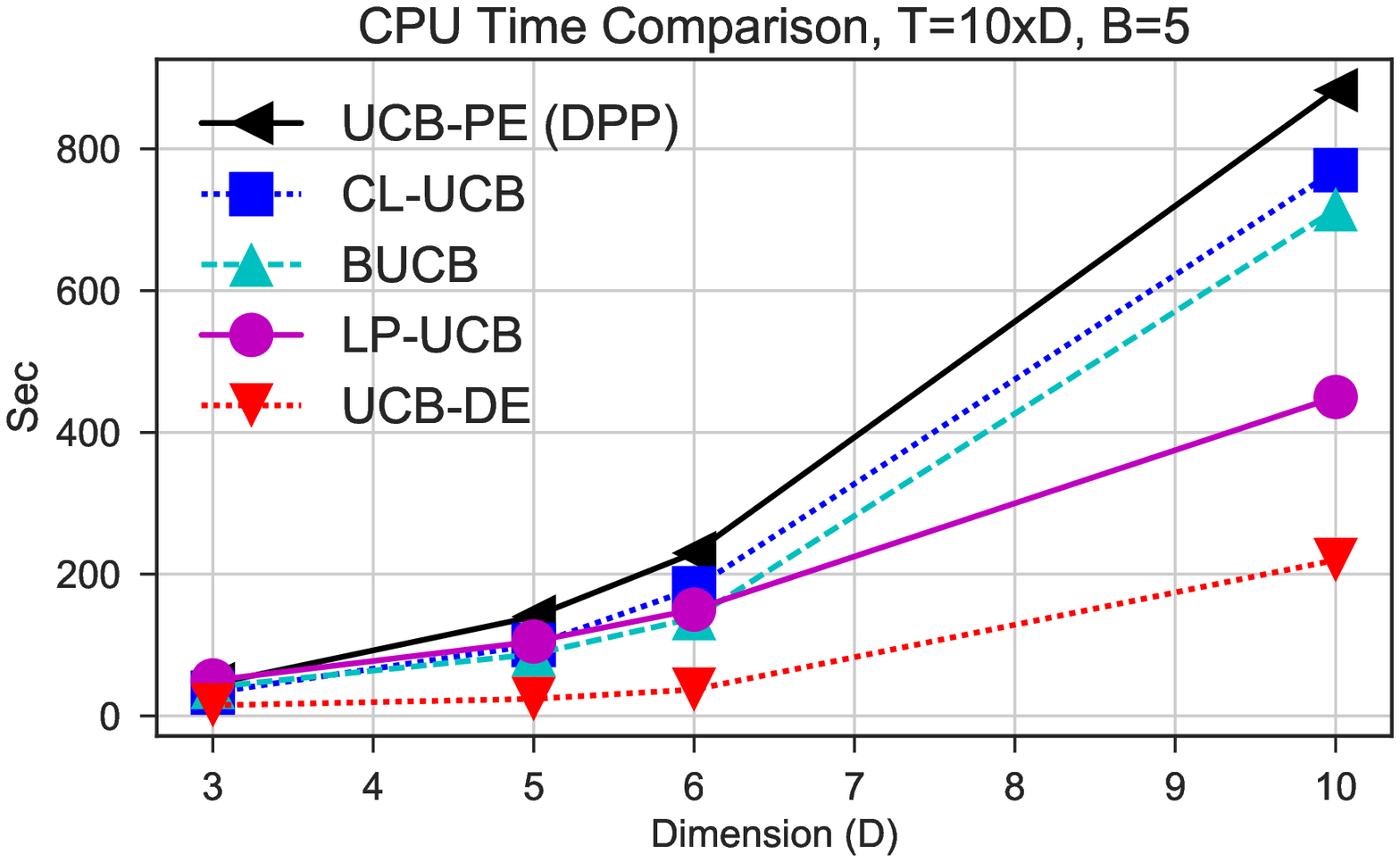}\caption{Time comparison with different batch size $B$ and dimensions $d$.
The proposed UCB-DE is the fastest batch BO method.\label{fig:Batch-BO-time} }
\par\end{centering}
\vspace{-0pt}
\end{figure}

\subsection{Computation complexity comparison}

In Fig. \ref{fig:Batch-BO-time}, we study the computational time
w.r.t. different batch sizes $B=5,10,15,20$ on Hartmann 6D function
and different dimensions $d=3,5,6,10$.

We learn from Fig. \ref{fig:Batch-BO-time} that UCB-PE (DPP) takes
more computation than the others because it repeatedly computes relevant
regions and searches for the highest variance location in these regions.
BUCB only updates the predictive variance while it keeps the mean
function fixed. Hence, BUCB is cheaper than CL which requires updating
both mean and variance. Local penalization (LP) \cite{Gonzalez_2015Batch}
is faster than BUCB. This is because LP does not recompute GP after
adding each element in a batch, instead LP maintains a penalized cost
after inserting each element. 

Our proposed approach significantly runs faster than the others. Especially,
when a batch size $B$ increases, our UCB-DE computation seems invariant
and surpasses the others in an order of magnitude. Because increasing
a batch size results in more observations and requires to perform
more global optimization steps, the complexity of BUCB is increased
while our UCB-DE is much cheaper.

%% file: Conclusion.tex
In this paper, we have considered a novel setting in optimizing the
less expensive black-box functions. To address this less expensive
setting, we have presented an algorithm UCB-DE for batch Bayesian
optimization. While our approach is simple to implement, it maintains
desirable properties of exploration. Our proposed DE greatly speeds
up the computation for batch BO that we do not need to run multiple
global optimization to fill a batch. We demonstrate that the proposed
UCB-DE is the best batch BO approach for optimizing the less expensive
black-box functions.